# Realizing an Optimization Approach Inspired from Piaget's Theory on Cognitive Development

*Utku Kose*
Computer Sciences Application and Research Center
Usak University, Usak, Turkey
E-mail: utku.kose@usak.edu.tr

*Ahmet Arslan*
Department of Computer Engineering
Selcuk University, Konya, Turkey
E-mail: ahmetarslan@selcuk.edu.tr

**Abstract**
The objective of this paper is to introduce an artificial intelligence based optimization approach, which is inspired from Piaget's theory on cognitive development. The approach has been designed according to essential processes that an individual may experience while learning something new or improving his / her knowledge. These processes are associated with the Piaget's ideas on an individual's cognitive development. The approach expressed in this paper is a simple algorithm employing swarm intelligence oriented tasks in order to overcome single-objective optimization problems. For evaluating effectiveness of this early version of the algorithm, test operations have been done via some benchmark functions. The obtained results show that the approach / algorithm can be an alternative to the literature in terms of single-objective optimization. The authors have suggested the name: Cognitive Development Optimization Algorithm (CoDOA) for the related intelligent optimization approach.

## 1. Introduction

Nowadays, developments and improvements in different technologies have active role on ensuring better life standards. It is also clear that newer solutions, which are used effectively for real-world problems, are widely designed thanks to technological developments and improvements. On the other hand, multidisciplinary interactions also have an important role on designing new solution approaches, methods, or techniques for specific fields. When we focus more on recent scientific works, it can be seen that theories, designed mathematical approaches are employed within very different fields in order to overcome some problems or try to obtain alternative solution ways. Another important point is that such interactions are because of common problems that are solved in the context of different fields. For example, the problem of optimization has been widely taken into consideration by researchers although their interest fields are different.

If we examine different literatures associated with employment of optimization problems, we can figure out that there is a remarkable research interest in solving optimization problems by using alternative ways. Recently, there has been a great momentum in designing novel methods, and techniques as alternative solution ways to real-world based optimization problems. At this point, also some specific interest fields have appeared as a result of many different research works reported. Swarm intelligence is one of these fields and it is an important focus point within which effective and accurate solutions for optimization tasks are introduced, thanks to strong relations with different disciplines and bio-inspired computation (Blum & Li, 2008), (Garnier, Gautrais & Theraulaz, 2007), (Hinchey, Sterritt & Rouff, 2007), (Kennedy, Kennedy, Eberhart & Shi, 2001), (Parpinelli & Lopes, 2011), (Yang, Cui, Xiao, Gandomi & Karamanoglu, 2013). Swarm intelligence is mainly related to the scope of artificial intelligence and because of this; especially intelligent optimization solution approaches, methods, or techniques are often developed. Intelligent algorithms, designed and developed in the context of swarm intelligence employs some essential features and functions, which are inspired from swarms like fish swarms, bird swarms in the nature. Additionally, such features and functions are used for obtaining new swarm intelligence oriented algorithms, which are focused on some theories, mathematical systems…etc. from different fields. Artificial Intelligence is the science and research field of the future. Furthermore, it can be expressed that swarm intelligence has an important potential on shaping the solution ways of the future; as an active sub-field of artificial intelligence.

This paper introduces an artificial intelligence based optimization approach, which is inspired from Piaget's theory on cognitive development. Briefly, the approach has been designed according to essential processes that an individual may experience while learning something new or improving his / her knowledge. These processes are associated with the Piaget's ideas on an individual's cognitive development. Generally, the approach explained within this paper is a simple algorithm, which includes swarm intelligence oriented tasks in order to overcome single-objective optimization problems. It has been aimed to form an easy-to-apply algorithmic form in order to understand its solution approach easily and make it portable to integrate any problem. In order to evaluate the effectiveness of this early version of the algorithm, test operations have been done via some benchmark functions. The authors have suggested the name: Cognitive Development Optimization Algorithm (CoDOA) for this intelligent optimization approach.

According to the objective of this paper, the remaining content is organized as follows: The next section provides a look to the history of the approach and the development process so far. Following to that section, the third section expresses the fundamentals of the developed algorithm. It briefly explains the approach and provides steps of the designed algorithm, which is named as the Cognitive Development Optimization Algorithm (CoDOA). Next to the third section, a brief evaluation process performed via some optimization benchmark functions is reported under the fourth section. Finally, the paper is ended by discussing about conclusions and future works.

## 2. The History of the Optimization Approach and Development

Just before focusing on the optimization approach inspired from Piaget's theory on cognitive development, it is a good idea to focus on the history of the approach and development of the algorithm. In this sense, the foremost points regarding to 'the history' can be explained as follows:

- While working on developments regarding to another artificial intelligence based optimization algorithm: Vortex Optimization Algorithm (VOA), which has been developed by the same authors, a new idea on using essentials of Piaget's related theory to develop alternative approach has been thought.

- After first ideas, the authors have focused on Piaget's theory on cognitive development; in order to understand more about this theory and its detail.

- After having enough knowledge on 'Piaget's theory', it has been tried to form connections between theoretical aspects of the theory and essential features / functions often employed within swarm intelligence.

- While designing the algorithm structure, it has been aimed to use simple equations as similar to the approach used in development process of the VOA. It is also important that the authors have inspired from essential equations used along the VOA; in order to ensure simple swarm movements towards better values.

- While forming algorithmic connections to the Piaget's theory on cognitive development, the authors have done some experimental studies regarding to usage of additional parameters along the new approach. Finally, some parameters like interactivity rate or maturity limit have been included in the related algorithm structure.

After providing a brief report on history of the approach and so development process, it is now better to explain details of the algorithmic structure.

### 3. An A.I. Based Optimization Approach Inspired from Piaget's Theory on Cognitive Development

In general, Cognitive Development Optimization Algorithm (CoDOA) is an artificial intelligence based, single-objective optimization approach, which comes with simple equations and inspires from Piaget's theory on cognitive development. Cognitive development can be defined briefly as a natural development process related to each individual. Piaget has expressed that individuals generally have different stages like maturation, social interaction, balancing while learning new concepts and improving his / her cognitive infrastructure (Piaget, 1964, 1973), (Singer & Revenson, 1997).

CoDOA employs steps regarding to following phases: Initialization Phase, Socialization Phase, Maturation Phase, Rationalizing Phase, and Balancing Phase. The related phases are some kind of calculation steps, which have been formed as a result of inspirations from stages of cognitive development. The phases are repeated until the stopping criterion is met.

Algorithmic details of the CoDOA can be expressed briefly as follows:

- **Step 1 (Initialization Phase):** Set initial parameters (*N*: number of particles; initial interactivity rate (*ir*) and experience (*ex*) values for each particle; max. and min. limits (min. limit is 0.0) for *ir* value (*max_ir* and *min_ir*); *ml* for the maturity limit; and *r* for the rationality rate.

  Also, set other values related to the function, problem…etc. (e.g. dimension, search domain…etc.).

- **Step 2:** Place the particles randomly in the solution space, and calculate fitness values for each of them. Update the *ir* value of the particle with the best fitness value by using a random value (Equation 1).

  $$best\_particle\_ir\_(new) = best\_particle\_ir\_(current) + (rand. * best\_particle\_ir\_(current)) \quad (1)$$

  Also, increase *ex* value of this particle by 1.

- **Step 3:** Repeat the loop steps below until the stopping criterion (e.g. iteration number) is met:
  - **Step 3.1 (Socialization Phase):** Decrease (by 1) *ex* value of each particle, whose fitness value is equal to or above the average fitness of all particles (if the problem is minimization).

    Also, increase (by 1) *ex* value of each particle, whose fitness value is under the average fitness of all particles (if the problem is minimization). Finally, Update the *ir* value of these particles by using a random value (Equation 2).

    $$particle_j\_ir\_(new) = particle_j\_ir\_(current) + (rand. * particle_j\_ir\_(current)) \quad (2)$$

  - **Step 3.2:** Update the *ir* value of all particles by using the Equation 3:

    $$particle_i\_ir\_(new) = rand. * particle_i\_ir\_(current) \quad (3)$$

17

- **Step 3.3:** Update position of each particle (except from the best particle so far) by using the Equation 4:

$$particle_i\_pos.\_(new) = particle_i\_pos.\_(current) + (rand. * (particle_i\_ir\_(current) * (global\_best\_pos. - particle_i\_pos.\_(current)))) \quad (4)$$

- **Step 3.4:** Calculate fitness values according to new positions of each particle. Update the *ir* value of the particle with the better / best fitness value by using a random value (Equation 5).

$$best\_particle\_ir\_(new) = best\_particle\_ir\_(current) + (rand. * best\_particle\_ir\_(current)) \quad (5)$$

Also, increase ex value of this particle by 1.

- **Step 3.5 (Maturation Phase):** Update *ir* value of each particle, whose *ex* value is equal to or under the *ml* value by using the Equation 6:

$$particle_j\_ir\_(new) = particle_j\_ir\_(current) + (rand. * particle_j\_ir\_(current)) \quad (6)$$

Calculate fitness values according to new positions of each particle. Update the *ir* value of the particle with the better / best fitness value by using a random value (Equation 7).

$$best\_particle\_ir\_(new) = best\_particle\_ir\_(current) + (rand. * best\_particle\_ir\_(current)) \quad (7)$$

Also, increase ex value of this particle by 1.

- **Step 3.6 (Rationalizing Phase):** Update *ir* and positions of each particle, whose *ex* value is under 0, by using the following equations:

$$particle_j\_ir\_(new) = particle_j\_ir\_(current) + (rand. * (best\_particle\_ir\_(current) / particle_j\_ir\_(current))) \quad (8)$$

$$particle_i\_pos.\_(new) = particle_i\_pos.\_(current) + (rand. * (particle_i\_ir\_(current) * (global\_best\_pos. - particle_i\_pos.\_(current)))) \quad (9)$$

Update *ir* of each particle, whose *ex* value is equal to or above 0, and repeat this *r* times; by using the Equation 10:

$$particle_j\_ir\_(new) = particle_j\_ir\_(current) + (rand. * (best\_particle\_ir\_(current) / particle_j\_ir\_(current))) \quad (10)$$

- **Step 3.7 (Balancing Phase):** Update the *ir* value of all particles by using the Equation 11:

$$particle_i\_ir\_(new) = rand. * particle_i\_ir\_(current) \quad (11)$$

Calculate fitness values according to new positions of each particle. Update the *ir* value of the particle with the better / best fitness value by using a random value (Equation 12).

$$best\_particle\_ir\_(new) = best\_particle\_ir\_(current) + (rand. * best\_particle\_ir\_(current)) \quad (12)$$

    Also, increase *ex* value of this particle by 1. Return to the Step 3.1. if the stopping criteria is not achieved yet.

- **Step 4:** The best values obtained within the loop are related to the optimum solution.

A flowchart regarding to the CoDOA is shown in Figure 1.

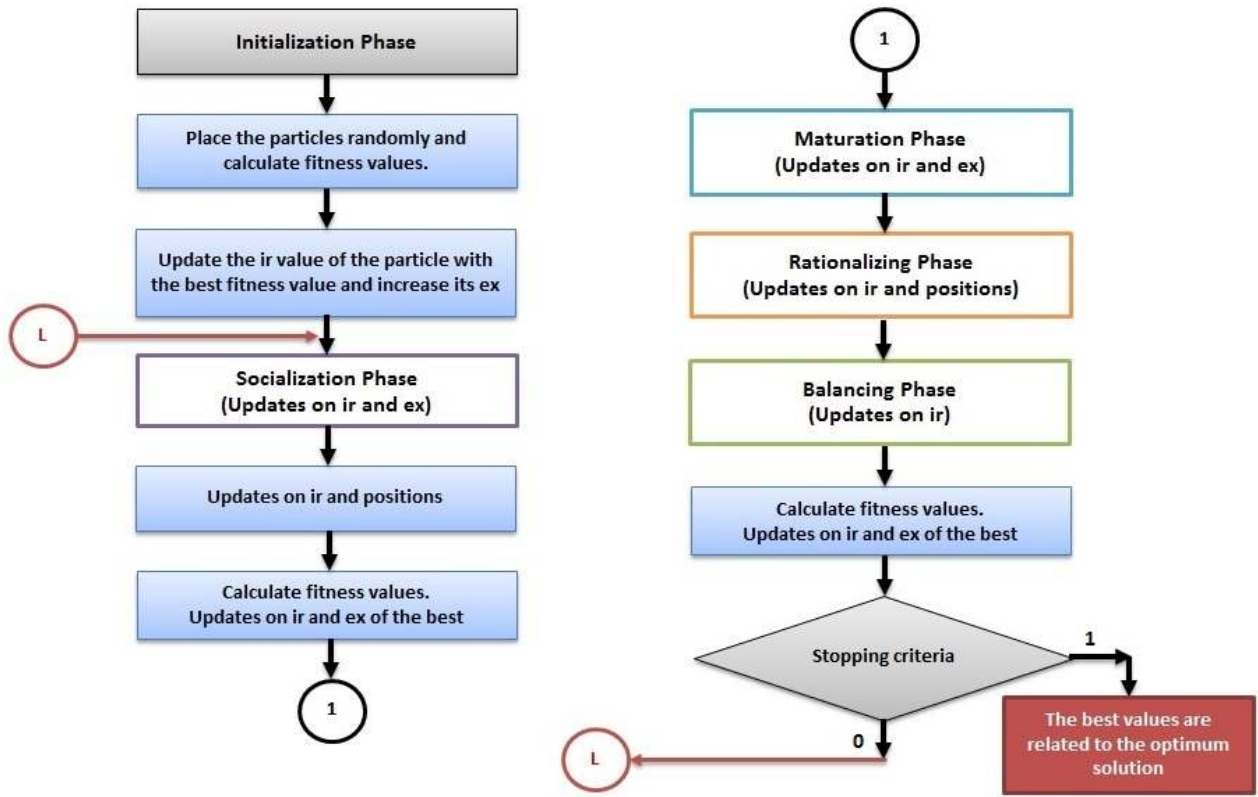

*Figure 1.Flowchart of the CoDOA.*

### 4. Evaluation

For having an idea about the effectiveness of the CoDOA, some single-objective optimization benchmark functions have been solved in terms of different dimensions. General conditions for the performed tests are as follows:

- Total number of particles (*N*): 50
- Total iteration (the stopping criteria): 5000
- Initial interactivity rate: 0.50

- Max. interactivity rate: 10.0
- Maturity limit (*ml*): 3
- Rationality rate (*r*): 2

- Dimensions: 2 for first four functions and 2, 5, 10, 20, and 30 respectively for the last two functions.
- Remaining specific values have been determined according to characteristics of each function.

Table 1 shows brief information regarding to the benchmark functions used along the test process.

Table 1. Benchmark functions used along the test process.

| Function | Formula | Search Domain | Minimum |
|---|---|---|---|
| Booth's | $f(x,y) = (x + 2y - 7)^2 + (2x + y - 5)^2$ | $-10 \leq x, y \leq 10$ | $f(1,3) = 0$ |
| Beale's | $f(x,y) = (1.5 - x + xy)^2 + (2.25 - x + xy^2)^2 + (2.625 - x + xy^3)^2$ | $-4.5 \leq x, y \leq 4.5$ | $f(3, 0.5) = 0$ |
| Goldstein–Price | $f(x,y) = \bigl(1 + (x+y+1)^2(19 - 14x + 3x^2 - 14y + 6xy + 3y^2)\bigr)\bigl(30 + (2x - 3y)^2(18 - 32x + 12x^2 + 48y - 36xy + 27y^2)\bigr)$ | $-2 \leq x, y \leq 2$ | $f(0, -1) = 3$ |
| McCormick | $f(x,y) = \sin(x + y) + (x - y)^2 - 1.5x + 2.5y + 1$ | $-1.5 \leq x \leq 4$ $-3 \leq y \leq 4$ | $f(-0.54719, -1.54719) = -1.9133$ |
| Three-hump camel | $f(x,y) = 2x^2 - 1.05x^4 + \dfrac{x^6}{6} + xy + y^2$ | $-5 \leq x, y \leq 5$ | $f(0, 0) = 0$ |
| Sphere | $f(x) = \sum_{i=1}^{n} x_i^2$ | $-100 \leq x_i \leq 100$ $1 \leq i \leq n$ | $f(x_1, \ldots, x_n) = f(0, \ldots, 0) = 0$ |
| Rosenbrock | $f(x) = \sum_{i=1}^{n-1} [100(x_{i+1} - x_i^2)^2 + (x_i - 1)^2]$ | $-30 \leq x_i \leq 30$ $1 \leq i \leq n$ | $f(x_1, \ldots, x_n) = f(1, \ldots, 1) = 0$ |

Table 2 provides a brief report for the obtained results after the test operations done with benchmark functions.

Table 2. Obtained results after the test operations done with benchmark functions.

| Function | Minimization Results for the Dimensions as: | | | | |
|---|---|---|---|---|---|
| | 2 | 5 | 10 | 20 | 30 |
| Booth's | 0.0000 | x | x | x | x |
| Beale's | 0.0000 | x | x | x | x |
| Goldstein–Price | 3.0000 | x | x | x | x |
| McCormick | -1.9133 | x | x | x | x |
| Three-hump camel | 0.0000 | x | x | x | x |
| Sphere | 0.0000 | 0.0000 | 0.0000 | 0.0000 | 0.0000 |
| Rosenbrock | 0.0000 | 0.0000 | 0.0000 | 0.0161 | 8.1197 |
| x: not applicable | | | | | |

According to the Table 2, the CoDOA reaches to the desired values for the used benchmark functions. Except from the Rosenbrock with 30 dimensions, CoDOA can be closer to the minimum values especially for bigger dimensional Sphere and Rosenbrock functions, although the number of total iteration and also of particles were low (5000 and 50).

**5. Conclusion and Future Work**

In this paper a new artificial intelligence based optimization approach, which is inspired from Piaget's theory on cognitive development, has been introduced. Called as the Cognitive Development Optimization Algorithm (CoDOA), the paper has provided information about how this approach has been realized in the context of an algorithmic structure and reported brief test / evaluation processes done via some benchmark functions. Generally, the CoDOA seems to effective enough for solving single-objective optimization problems. In the sense of the performed evaluation process, it may need additional modifications (or new tests with changed initial parameters) only for Rosenbrock function with dimension higher than 20.

In the sense of the CoDOA, some future works can be explained briefly as follows:
- Further works will be based on also comparison of the CoDOA with other alternative algorithms in the literature.
- This paper has focused on some single-objective benchmark functions. So, there can be more works on other remaining benchmark functions.
- Because of results on Rosenbrock function with dimension of 30, there will be additional works to see if algorithm can be improved or the current early version can be adapted with changed initial parameters.
- The authors have already started some works regarding to optimization problems from different fields.
- There will be also some more works on CoDOA particle / algorithm parameters; in order see if changed parameters can improve performance and accurateness.

**References**

Blum, C., & Li, X. (2008). Swarm intelligence in optimization (pp. 43-85). Springer Berlin Heidelberg.

Garnier, S., Gautrais, J., &Theraulaz, G. (2007). The biological principles of swarm intelligence. Swarm Intelligence, 1(1), 3-31.

Hinchey, M. G., Sterritt, R., &Rouff, C. (2007). Swarms and swarm intelligence. Computer, 40(4), 111-113.

Kennedy, J., Kennedy, J. F., Eberhart, R. C., & Shi, Y. (2001). Swarm intelligence. Morgan Kaufmann.

Parpinelli, R. S., & Lopes, H. S. (2011). New inspirations in swarm intelligence: a survey. International Journal of Bio-Inspired Computation, 3(1), 1-16.

Piaget, J. (1964). Part I: Cognitive development in children: Piaget development and learning, Journal of Research in Science Teaching, 2(3), 176-186.

Piaget, J. (1973). Main Trends in Psychology. London: George Allen &Unwin.

Singer, D. G. & Revenson, T. A. (1997). A Piaget primer: How a child thinks. International Universities Press, Inc., 59 Boston Post Road, Madison, CT 06443-1524.

Yang, X. S., Cui, Z., Xiao, R., Gandomi, A. H., &Karamanoglu, M. (Eds.). (2013). Swarm intelligence and bio-inspired computation: theory and applications. Newnes.